\documentclass[letterpaper]{article} 
\usepackage{aaai2026}
\usepackage{times}  
\usepackage{helvet}  
\usepackage{courier}  
\usepackage[hyphens]{url}  
\usepackage{graphicx} 
\usepackage{natbib}  
\usepackage{caption} 
\usepackage{algorithm}
\usepackage{algorithmic}

\usepackage{newfloat}
\usepackage{listings}

\DeclareCaptionStyle{ruled}{labelfont=normalfont,labelsep=colon,strut=off} 
\floatstyle{ruled}
\newfloat{listing}{tb}{lst}{}
\floatname{listing}{Listing}
\nocopyright 

\title{Refine-and-Contrast: Adaptive Instance-Aware BEV Representations for Multi-UAV Collaborative Object Detection}
\author {
    Zhongyao Li\textsuperscript{\rm 1, \rm 2},
    Peirui Cheng\textsuperscript{\rm 1},
    Liangjin Zhao\textsuperscript{\rm 1},
    Chen Chen\textsuperscript{\rm 1, \rm 2},
    Yundu Li\textsuperscript{\rm 1, \rm 2},
    Zhechao Wang\textsuperscript{\rm 1, \rm 2},
    Xue Yang\textsuperscript{\rm 3},
    Xian Sun\textsuperscript{\rm 1},
    Zhirui Wang\textsuperscript{\rm 1}\thanks{Corresponding author.}
}
\affiliations {
    \textsuperscript{\rm 1}Key Laboratory of Target Cognition and Application Technology, \\ Aerospace Information Research Institute, Chinese Academy of Sciences\\
    \textsuperscript{\rm 2}University of Chinese Academy of Sciences \ 
    \textsuperscript{\rm 3}Shanghai Jiao Tong University
}

\usepackage{amssymb}
\usepackage{amsmath}
\usepackage{multirow}
\usepackage{makecell}
\usepackage{booktabs}

\begin{document}

\maketitle

\begin{abstract}
Multi-UAV collaborative 3D detection enables accurate and robust perception by fusing multi-view observations from aerial platforms, offering significant advantages in coverage and occlusion handling, while posing new challenges for computation on resource-constrained UAV platforms. In this paper, we present AdaBEV, a novel framework that learns adaptive instance-aware BEV representations through a refine-and-contrast paradigm. Unlike existing methods that treat all BEV grids equally, AdaBEV introduces a Box-Guided Refinement Module (BG-RM) and an Instance-Background Contrastive Learning (IBCL) to enhance semantic awareness and feature discriminability. BG-RM refines only BEV grids associated with foreground instances using 2D supervision and spatial subdivision, while IBCL promotes stronger separation between foreground and background features via contrastive learning in BEV space. Extensive experiments on the Air-Co-Pred dataset demonstrate that AdaBEV achieves superior accuracy-computation trade-offs across model scales, outperforming other state-of-the-art methods at low resolutions and approaching upper bound performance while maintaining low-resolution BEV inputs and negligible overhead.
\end{abstract}


\section{Introduction}

\quad Multi-UAV collaborative 3D detection~\cite{zhu2020multi, hu2022where2comm, tian2024ucdnet} aims to fuse multi-view observations from multiple UAVs to achieve more comprehensive and accurate object perception. Compared to single-UAV setups, collaborative detection significantly improves perception range, occlusion mitigation, and detection robustness. These advantages are particularly crucial in real-world applications such as surveillance~\cite{chang2018surveillance}, urban monitoring~\cite{nguyen2021drone}, and disaster response~\cite{qu2023environmentally}, where timely and reliable aerial 3D understanding is essential for efficient multi-agent coordination.


For collaborative 3D detection tasks of multi-UAVs, the Bird's-Eye View (BEV) representation serves as a crucial intermediate format by projecting 3D spatial information onto a unified top-down plane, thus allowing effective fusion and alignment of perception data from multiple points of view~\cite{philion2020lift}. As a result, BEV has become a widely adopted representation in multi-agent aerial systems. Current mainstream BEV-based 3D detection methods predominantly follow a Transformer-based paradigm~\cite{li2022bevformer, pan2024clip, yang2024widthformer}. These approaches typically initialize a set of spatially uniform grid queries in the BEV space and interact with multi-view sensory inputs (e.g., images, point clouds) alongside prior information, including camera poses. Through attention mechanisms~\cite{vaswani2017attention}, the model dynamically aggregates observations from different perspectives, facilitating unified object modeling and detection in the 3D space.

\begin{figure}[t]
\centering
\includegraphics[width=0.9\columnwidth]{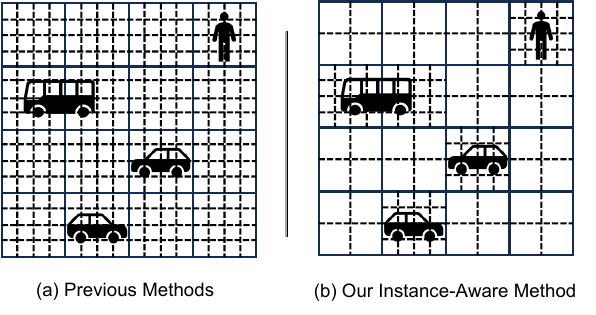} 
\caption{\textbf{Comparison between our and previous methods.} We break the conventional paradigm of \textit{uniform} BEV modeling by adaptively refining target regions to produce instance-aware BEV representations. Each grid represents a BEV query.
}
\label{fig0}
\end{figure}

However, we observe that most existing Transformer-based BEV representation approaches~\cite{li2022bevformer, yang2023bevformer, pan2024clip} operate under a \textit{uniform} assumption, treating all grids as equally important during the generation of BEV features, as shown in Figure~\ref{fig0}(a). This assumption stands in contrast to the inherently non-uniform information distribution observed from the perspective of UAVs, particularly on resource-constrained aerial platforms. Specifically, modeling all BEV regions equally disregards the semantic and detection significance of critical areas (e.g., near targets or prominent structural regions), leading to a mismatch between computational resource allocation and perceptual importance. This issue is further exacerbated by the unique characteristics of aerial viewpoints, where objects tend to be small in volume, sparsely distributed, and visually ambiguous at the boundaries. Such properties intensify the information imbalance between foreground and background regions in the BEV space, making the uniform treatment of BEV grids a critical bottleneck to detection performance.

To enable \textit{non-uniform} differentiated modeling, as illustrated in Figure~\ref{fig0}(b), we propose refining only the BEV grids corresponding to foreground regions, thereby enhancing the network’s capacity to model critical areas while avoiding redundant resource allocation to low-value background regions. Thus, we propose AdaBEV, a multi-UAV collaborative 3D detection model that adaptively distinguishes foreground from background in BEV representations. AdaBEV consists of two key components: the Box-Guided Refinement Module (BG-RM) and the Instance-Background Contrastive Learning (IBCL) module. BG-RM leverages 2D detection boxes to identify semantically salient regions and refines only those BEV grids whose projected spatial samples fall within the detected areas, using a local 4×4 subdivision to enhance spatial precision with minimal overhead. IBCL introduces a contrastive learning strategy directly in the BEV space, where instance-level foreground features (projected from ground-truth 3D boxes) are contrasted with randomly sampled background features to promote intra-class compactness and inter-class separability. These modules enable AdaBEV to move beyond simple task supervision and learn BEV features with stronger semantic discriminability and instance awareness.

Extensive experimental results on the Air-Co-Pred dataset~\cite{wang2024drones} demonstrate that our method achieves an excellent trade-off between accuracy and computation across different model scales. Under a low BEV resolution (e.g., $50\times50$), it significantly outperforms the BEVFormer~\cite{li2022bevformer} baseline (0.783 $\mathrm{vs.}$ 0.759) and approaches the upper bound performance achieved with higher resolutions (e.g., $200\times200$) with almost no additional computational overhead (141.56 $\mathrm{vs.}$ 141.49 GFLOPs), validating its effectiveness in achieving high precision with lightweight deployment.

\section{Related Work}

\subsubsection{Multi-UAV Collaborative Perception.} Multi-UAV collaborative perception has gained traction for enhancing situational awareness in complex environments. Recent methods adopt Bird's-Eye View (BEV) representations to enable spatial alignment and collaborative modeling in a unified top-down space. UCDNet~\cite{tian2024ucdnet} employs ground-plane depth priors and a geometric consistency loss to improve cross-view feature alignment. DHD~\cite{wang2024drones} integrates ground priors and height cues to construct BEV features and adaptively selects collaboration regions via a sliding-window mechanism for efficient trajectory forecasting.

While effective, these methods rely on accurate depth estimation, which remains challenging in aerial settings. To address this, we propose a BEV representation learning approach that accounts for the non-uniformity of feature contributions, enabling robust collaborative perception without precise depth supervision.

\begin{figure*}[t]
\centering
\includegraphics[width=1.0\textwidth]{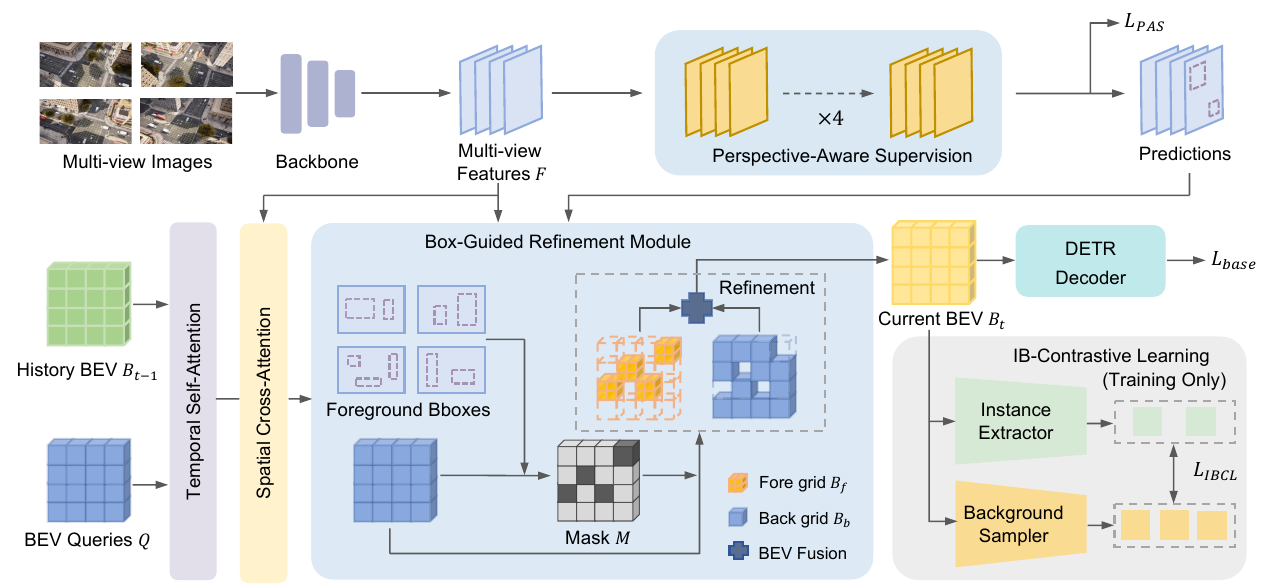} 
\caption{\textbf{Overall architecture of AdaBEV.} AdaBEV consists of two key components: a box-guided refinement module (BG-RM) that enhances foreground features using detections from a perspective-aware supervision (PAS) branch, and an instance-background contrastive learning (IBCL) module that enforces discriminative BEV representations via self-supervised learning.}
\label{fig1}
\end{figure*}

\subsubsection{BEV Generation.} Bird's-Eye View (BEV) representation~\cite{philion2020lift, liu2022petr, zhou2022cross,li2023fb} plays a key role in multi-agent perception due to its regular spatial layout, enabling efficient sensor fusion and spatial reasoning. Existing camera-only BEV generation methods fall into two categories: geometry-based and Transformer-based. Among geometry-based methods, Lift-Splat-Shoot~\cite{philion2020lift} lifts image features into frustum space via depth and projects them onto the BEV plane. Building on this idea, BEVDet~\cite{huang2021bevdet} further improves feature quality through temporal alignment, while BEVDet4D~\cite{huang2022bevdet4d} models spatio-temporal consistency for dynamic scenes. Transformer-based approaches, such as BEVFormer~\cite{li2022bevformer}, treat BEV generation as a sequence-to-sequence problem and aggregate multi-view features via attention without explicit depth. BEVFormerV2~\cite{yang2023bevformer} adopts a two-stage architecture, where proposals from the perspective head guide BEV queries in the second stage. 

Assuming uniform grid importance, existing methods often incur heavy computational costs from cross-view and high-resolution processing, limiting their scalability. In contrast, we model non-uniform, instance-aware BEV contributions to enable adaptive representation learning.

\subsubsection{Auxiliary Tasks for 3D Object Detection.} Auxiliary tasks have proven effective in improving feature quality and boosting 3D detection performance in surround-view settings. Focal-PETR~\cite{wang2023focal} introduces 2D detection supervision to guide 3D queries toward salient foreground regions. BEVFormerV2~\cite{yang2023bevformer} introduces an auxiliary 3D detection head from the perspective view to provide additional supervision. MV2D~\cite{wang2023object} treats 2D detections as sparse queries to constrain feature aggregation via sparse cross-attention. Far3D~\cite{jiang2024far3d} enhances long-range perception by constructing adaptive 3D queries from 2D boxes and depth predictions.

\section{Method}

\quad To address the limitations of existing BEV representation methods that rely on the assumption of spatial uniformity, we propose AdaBEV, an instance-aware framework for non-uniform BEV modeling. We begin by illustrating the overall architecture of AdaBEV in Section~\ref{sec:3.1}. Subsequently, we dive into the specifics of the BG-RM and IBCL modules in Section~\ref{sec:3.2} and Section~\ref{sec:3.3}, respectively. The detection decoder and the training loss of our framework are outlined in Section~\ref{sec:3.4}.

\subsection{Overall Architecture}
\label{sec:3.1}

\quad As shown in Figure~\ref{fig1}, AdaBEV builds on vanilla BEVFormer~\cite{li2022bevformer} with two core designs: a Box-Guided Refinement Module (BG-RM) that refines foreground regions based on the detected boxes generated from perspective-aware supervision (PAS) branch, and an Instance-Background Contrastive Learning (IBCL) that applies self-supervised contrastive learning directly in the BEV feature space.

During inference, multi-view RGB images are first fed into an image backbone to extract image features $F=\{F_i\}_{i=1}^{N_C}$, where $F_i$ denotes the view feature of $i$-th camera view, and $N_c$ is the total number of cameras. These image features are then passed through the perspective-aware supervision module to generate a set of bounding boxes $B=\{B_i\}_{i=1}^{N_C}$. We initialize BEV queries $Q\in \mathbb{R}^{H\times W\times C}$, where $H$ and $W$ define the BEV spatial resolution (e.g., $50\times50$), and $C$ is the channel dimension. Through temporal self-attention and spatial cross-attention, we obtain BEV feature $B_c\in \mathbb{R}^{H\times W\times C}$. Given the image features $F$, 2D bounding boxes $B$, and BEV features $B_c$, the BG-RM first generates a binary mask $M\in \mathbb{R}^{H\times W}$ via simple geometric projection. BEV grids whose projected points fall within the detection boxes are designated as foreground grids $B_f$ and are subjected to further refinement. The refined foreground grids $B_{f}^{'}$ are then fused with the unrefined background grids $B_b$ to produce the refined BEV representation $B_r\in \mathbb{R}^{H\times W\times C}$. After passing through $n$ encoder layers, we obtain the final BEV feature $B_t\in \mathbb{R}^{H\times W\times C}$, which is used for the subsequent 3D object detection task.

\subsection{Box-Guide Refinement Module}
\label{sec:3.2}

\begin{figure}[t]
\centering
\includegraphics[width=1.0\columnwidth]{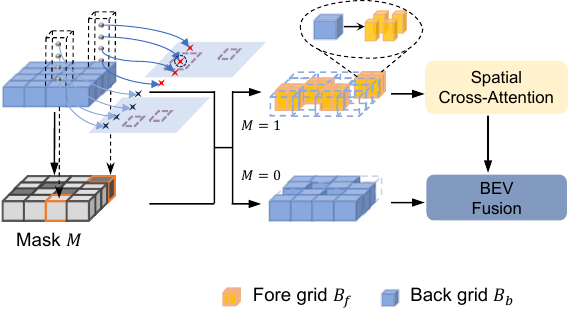} 
\caption{\textbf{Illustration of Box-Guided Refinement Module (BG-RM).} BEV grids with projected points inside detection boxes ($M=1$) are refined into high-resolution grids (orange) and process SCA recalculation, then fused with unrefined grids (blue) to produce enhanced BEV features.}
\label{fig2}
\end{figure}

\quad Guided by the prediction boxes in the image domain, BG-RM focuses on enhancing feature representations in the foreground-dense regions of the BEV space. As shown in Figure~\ref{fig2}, BG-RM first generates a foreground mask to locate informative regions, then applies refined spatial cross-attention over these regions, and finally fuses the refined foreground features with the background to obtain enhanced BEV representations.

\subsubsection{Foreground Mask Generation.}

To localize foreground regions in the BEV space, we first construct a binary mask $M\in \mathbb{R}^{H\times W}$ guided by predicted 2D boxes from the perspective-aware supervision (PAS) module. Specifically, at the location of each BEV grid $\mathbf{p}_{xy}$, we uniformly sample spatial points $z_{j}$ within its pillar and project them onto the image plane based on the camera’s intrinsic matrix $\mathbf{K}$, rotation matrix $\mathbf{R}$, and translation vector $\mathbf{t}$. If any of the projected points $\mathbf{u}_{xy}$ fall within a predicted bounding box from any camera view, the corresponding BEV grid is classified as foreground grid, and its mask value is set to 1. During training, we replace the predicted boxes with ground-truth annotations for more accurate supervision and stable optimization. This process can be expressed mathematically as:
\begin{equation}
{P}_{xy}=\left\{\mathbf{P}_{xy}^{j}=[\mathbf{p}_{xy},z_{j}]^{\top}\in\mathbb{R}^{3}\mid j=1,\ldots,N_{\mathrm{ref}}\right\},
\end{equation}
\begin{equation}
\mathbf{u}_{xy}^{j,i}=\mathbf{K}_{i}\left(\mathbf{R}_{i}\mathbf{P}_{xy}^{j}+\mathbf{t}_{i}\right),\quad i=1,\ldots,N_{c},
\end{equation}
\begin{equation}
M(x,y)=
\begin{cases}
1, & \!\!\!\begin{aligned}[t]
&\mathrm{if}\,\exists \, i\in\{1,\ldots,N_c\}, \\
&\exists \, j\in\{1,\ldots,N_{\mathrm{ref}}\},\ \mathbf{u}_{xy}^{j,i}\in B_i
\end{aligned} \\
0, & \mathrm{otherwise}
\end{cases},
\end{equation}
\noindent where $N_{\mathrm{ref}}$ represents the number of 3D reference points in one pillar.

\subsubsection{Refined Spatial Cross-Attention.}

In this work, the refinement strategy employed in BG-RM is built upon the Spatial Cross-Attention (SCA) originally proposed in BEVFormer~\cite{li2022bevformer}. The projection process of the original SCA can be formulated as:
\begin{equation}
\mathrm{SCA}(Q_{xy},F)=\sum_{i=1}^{N_{\mathrm{c}}}\sum_{j=1}^{N_{\mathrm{ref}}}\mathcal{F}_{d}(Q_{xy},{P}_{i}(x,y,z_{j}),F_{i}),
\end{equation}
\noindent for each point $(x,y)$ on the BEV plane, BEVFormer lifts it to $N_{\mathrm{ref}}$ 3D reference points by assigning different heights $z_j$. The deformable attention~\cite{zhu2020deformable} function $\mathcal{F}_{d}$ is then employed to sample features from the projected points $P_i(x,y,z_j)$ on the image features $F$, using $Q_{xy}$ as the query.

In BG-RM, local refinement is performed on BEV grid locations where the mask value is 1, i.e., those identified as foreground. Each foreground BEV grid is subdivided into $r\times r$ finer sub-grids (with $r=4$ by default), enabling higher spatial resolution. The partitioning of a sub-grid at location $(m, n)$ within a foreground grid can be formulated as:
\begin{equation}
x^{m} = x + \delta^{m}, \: y^{n} = y + \delta^{n}
\end{equation}
\noindent where $\delta$ denotes the offset of the sub-grid within the physical coordinate system relative to its parent BEV grid.

Cross-view feature interaction is then applied at sub-grid locations. Finally, the outputs of all $r\times r$ sub-grids are averaged to obtain the refined feature representation of the corresponding coarse grid. This process can be formulated as:
\begin{equation}
\mathrm{SCA}(Q_{xy}^{mn},F) = \sum_{i=1}^{N_c}\sum_{j=1}^{N_{\mathrm{ref}}}\mathcal{F}_d(Q_{xy},\mathcal{P}_i(x^{m},y^{n},z_j),F_i),
\end{equation}
\begin{equation}
\mathrm{SCA_r}(Q_{xy},F)=\frac{1}{r^2}\sum_{m,n=1}^r\mathrm{SCA}(Q_{xy}^{mn},F),
\end{equation}

\subsubsection{BEV Feature Fusion.}

We replace the original BEV queries at foreground locations with their refined features and retain the original queries for background regions. Compared to standard SCA, which uniformly models all BEV grids, our method selectively enhances the representational capacity for target areas while reducing redundant computation over the background regions.

\subsection{Instance-Background Contrastive Learning}
\label{sec:3.3}

\begin{figure}[t]
\centering
\includegraphics[width=1.0\columnwidth]{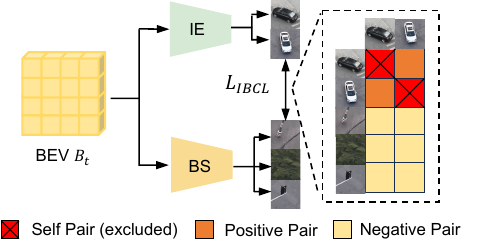} 
\caption{\textbf{Illustration of Instance-Background Contrastive Learning (IBCL).} The instance extractor (IE) projects ground-truth 3D boxes onto BEV to extract instance features, while the background sampler (BS) samples features from background regions. IBCL employs an InfoNCE loss to enforce instance-background feature separability.}
\label{fig3}
\end{figure}

\quad Existing BEV methods typically rely on indirect supervision via detection losses, which may cause task-specific overfitting and overlook the intrinsic structure of BEV features. To address this, IBCL introduces explicit contrastive supervision by enforcing semantic separation between instances and background. As shown in Figure~\ref{fig3}, IBCL extracts instance-level features via the instance extractor, samples distant background features using the background sampler, and applies an Instance-Background Contrastive Loss to enforce semantic separation in the BEV space.

\subsubsection{Instance Extractor.}

We begin by projecting each ground-truth bounding box $b_i$ onto the BEV space and cropping the corresponding region from the BEV feature $\mathbf{B}_{t}$. To account for spatial variations in semantic importance, we use a lightweight MLP to predict a spatial attention map over the adaptive pooled features, which are subsequently aggregated via weighted summation to produce the final instance feature $f_i$. The process can be formulated as:
\begin{equation}
\mathbf{w}_i=\mathrm{MLP}(\mathrm{AdaPool}(\mathrm{Crop}(\mathbf{B}_{t},b_i))),
\end{equation}
\begin{equation}
{f}_{i}=\text{WeightedPool}(\mathrm{AdaPool}(\mathrm{Crop}(\mathbf{B}_{t},b_{i})),\mathbf{w}_{i}),
\end{equation}

\subsubsection{Background Sampler.}

We perform random sampling within background regions and enforce a minimum distance constraint $d_{\mathrm{min}}$ from foreground region $\mathcal{R}_f$ to avoid semantic contamination. Sampled regions are processed with adaptive pooling, and the resulting features are aggregated via average pooling to yield the background representation $f_k$. The process can be formulated as:
\begin{equation}
\begin{aligned}
{f}_{k} &= \mathrm{AvgPool}(\mathrm{AdaPool}(\mathrm{Crop}(\mathbf{B}_{t},p_{k}))) \\
&\quad \mathrm{s.t.}\quad \mathrm{dist}(p_{k},\mathcal{R}_{f})>d_{\mathrm{min}},
\end{aligned}
\end{equation}

\subsubsection{Instance-Background Contrastive Loss.}

Inspired by image-level contrastive learning approaches such as SimCLR~\cite{chen2020simple}, we introduce a contrastive learning paradigm into the BEV space to construct a structured and discriminative supervisory signal. Instance and background features are projected into a shared embedding space and $l_2$-normalized to ensure consistent scaling for semantic contrast. Cosine similarity is then used to form a global similarity matrix for contrastive pair construction.

Built upon the InfoNCE principle~\cite{oord2018representation}, IBCL simultaneously enforces: (1) \textbf{inter-instance compactness}, promoting semantic coherence among instances; and (2) \textbf{instance-background separability}, discouraging representational entanglement between instance and background features. To the best of our knowledge, it is the first to explicitly model instance-level semantic distributions between foreground and background regions from a BEV perspective. The overall loss function can be formulated as:
\begin{equation}
\mathcal{L}_{IBCL}=-\frac{1}{|\mathcal{P}|}\sum_{(i,j)\in\mathcal{P}}\log\frac{\exp(\mathcal{S}_{ij})}{\exp(\mathcal{S}_{ij})+\sum_{k\in\mathcal{N}}\exp(\mathcal{S}_{ik})},
\end{equation}

\noindent where $\mathcal{P}$ and $\mathcal{N}$ represents the sets of positive samples and negative samples, respectively. $\mathcal{S}_{ij}=\cos{(f_i,f_j)}/\tau$ is the sample-wise similarity between $i$ and $j$. The temperature coefficient is denoted by $\tau$.

\begin{table*}[!t]
\fontsize{10}{11}\selectfont
\centering
\begin{tabular}{ccccccccccccccccccc}

\toprule
Backbone & \makecell[l]{Method} & BEV Resolution & AP\ensuremath{\uparrow} & ATE\ensuremath{\downarrow} & ASE\ensuremath{\downarrow} & AOE\ensuremath{\downarrow} \\

\midrule
\multirow{9}{*}{ResNet-50} & \makecell[l]{BEVDet~\cite{huang2021bevdet}} & $128\times128$ & 0.639 & 0.432 & 0.204 & 0.108 \\
& \makecell[l]{BEVDet4D~\cite{huang2022bevdet4d}} & $128\times128$ & 0.617 & 0.431 & 0.224 & 0.139 \\
& \makecell[l]{BEVLongTerm~\cite{huang2022bevdet4d}} & $128\times128$ & 0.697 & 0.330 & 0.186 & 0.129 \\
& \makecell[l]{BEVDet4D-Stereo~\cite{huang2022bevdet4d}} & $128\times128$ & 0.717 & 0.306 & 0.204 & 0.081 \\
& \makecell[l]{BEVDepth~\cite{li2023bevdepth}} & $128\times128$ & 0.721 & \textbf{0.295} & 0.186 & 0.075 \\
& \makecell[l]{UCDNet~\cite{tian2024ucdnet}} & $128\times128$ & 0.734 & 0.323 & 0.182 & 0.051 \\
& \makecell[l]{BEVFormer*~\cite{li2022bevformer}} & $50\times50$ & 0.759 & 0.407 & 0.149 & 0.045 \\
& \makecell[l]{AdaBEV (ours)} & $50\times50$ & 0.783 & 0.390 & 0.138 & 0.032 \\
& \makecell[l]{AdaBEV (ours)} & $128\times128$ & \textbf{0.790} & 0.380 & \textbf{0.132} & \textbf{0.029} \\

\midrule
\multirow{6}{*}{ResNet-101} & \makecell[l]{BEVDet~\cite{huang2021bevdet}} & $128\times128$ & 0.860 & 0.269 & 0.092 & 0.041 \\
& \makecell[l]{BEVDet4D~\cite{huang2022bevdet4d}} & $128\times128$ & 0.829 & 0.311 & 0.139 & 0.055 \\
& \makecell[l]{BEVDet4D-Stereo~\cite{huang2022bevdet4d}} & $128\times128$ & 0.905 & 0.220 & 0.135 & 0.050 \\
& \makecell[l]{BEVDepth~\cite{li2023bevdepth}} & $128\times128$ & 0.898 & \textbf{0.208} & 0.128 & 0.070 \\
& \makecell[l]{BEVFormer*~\cite{li2022bevformer}} & $150\times150$ & 0.904 & 0.257 & \textbf{0.068} & 0.033 \\
& \makecell[l]{AdaBEV (ours)} & $150\times150$ & \textbf{0.910} & 0.239 & \textbf{0.068} & \textbf{0.024} \\

\bottomrule

\end{tabular}
\caption{\label{tab:main}\textbf{3D Detection results on the Air-Co-Pred} $\mathrm{val}$ \textbf{set.} Comparison of 3D object detection performance across various BEV detectors. AdaBEV is applied to both the tiny and small variants of BEVFormer, and demonstrates consistent performance improvements compared to the corresponding baselines. *:Baseline methods for a fair comparison. \textbf{Bold}: Best. }
 
\end{table*}
\begin{table*}[!t]
\fontsize{10}{11}\selectfont
\centering
\begin{tabular}{cccccccccccccc}
\toprule
Method & BEV Resolution & Input Size & AP$\uparrow$ & ATE$\downarrow$ & ASE$\downarrow$ & AOE$\downarrow$ & GFLOPs & Param. \\

\midrule 
BEVFormer-tiny (lower) & $50\times50$ & (450, 800) & 0.759 & 0.407 & 0.149 & 0.045 & 141.49 & 33.57M \\ 
BEVFormer-tiny (upper) & $200\times200$ & (450, 800) & 0.786 & 0.386 & 0.142 & 0.031 & 364.06 & 52.80M \\
AdaBEV-R (ours) & $50\times50$ & (450, 800) & 0.775 & 0.395 & 0.140 & 0.032 &141.56 & 41.01M \\

\midrule
BEVFormer-small (lower) & $50\times50$ & (720, 1280) & 0.898 & 0.255 & 0.089 & 0.030 & 386.31 & 54.42M \\ 
BEVFormer-small (upper) & $200\times200$ & (720, 1280) & 0.912 & 0.219 & 0.089 & 0.029 & 608.88 & 64.06M \\ 
AdaBEV-R (ours) & $50\times50$ & (720, 1280) & 0.902 & 0.253 & 0.066 & 0.024 & 386.49 & 61.81M \\

\midrule
BEVFormer-base (lower) & $50\times50$ & (900, 1600) & 0.930 & 0.202 & 0.071 & 0.025 & 880.67 & 59.50M \\ 
BEVFormer-base (upper) & $200\times200$ & (900, 1600) & 0.940 & 0.164 & 0.075 & 0.023 & 1266.82 & 69.14M \\ 
AdaBEV-R (ours) & $50\times50$ & (900, 1600) & 0.934 & 0.196 & 0.071 & 0.025 & 886.32 & 66.89M \\

\bottomrule

\end{tabular}
\caption{\label{tab:up&low}\textbf{Performance and computation comparison under different BEV resolutions and model scales.} “AdaBEV-R” denotes a variant of our method trained without the IBCL supervision, in order to ensure a fair comparison with the baseline.
}
\end{table*}
\begin{table}[!t]
\fontsize{10}{11}\selectfont
\centering
\begin{tabular}{cccccc}
\toprule
\#UAVs & \makecell[l]{IDs} & AP$\uparrow$ & ATE$\downarrow$ & ASE$\downarrow$ & AOE$\downarrow$ \\

\midrule
2 & \makecell[l]{UAV\textsubscript{12}} & 0.556 & 0.549 & 0.177 & 0.049 \\
2 & \makecell[l]{UAV\textsubscript{13}} & 0.595 & 0.577 & 0.179 & 0.040 \\
3 & \makecell[l]{UAV\textsubscript{123}} & 0.676 & 0.477 & 0.161 & 0.047 \\
4 & \makecell[l]{UAV\textsubscript{1234}} & \textbf{0.783} & \textbf{0.390} & \textbf{0.138} & \textbf{0.032} \\
\bottomrule

\end{tabular}
\caption{\label{tab:collab}\textbf{Performance comparison under varying numbers of collaborative UAVs.} UAV\textsubscript{1} and UAV\textsubscript{2} are adjacent, while UAV\textsubscript{1} and UAV\textsubscript{3} are positioned diagonally.}
\end{table}
\begin{table}[!t]
\fontsize{10}{11}\selectfont
\centering
\begin{tabular}{c@{\hspace{0.7em}}c@{\hspace{0.7em}}c@{\hspace{1em}}cccc@{}}
\toprule
PAS & BG-RM & IBCL & AP$\uparrow$ & ATE$\downarrow$ & ASE$\downarrow$ & AOE$\downarrow$ \\

\midrule 
 - & - & - & 0.759 & 0.407 & 0.149 & 0.045 \\ 
\checkmark & - & - & 0.764 & 0.412 & 0.144 & 0.035 \\
\checkmark & \checkmark & - & 0.775 & 0.395 & 0.140 & \textbf{0.032} \\
\checkmark & \checkmark & \checkmark & \textbf{0.783} & \textbf{0.390} & \textbf{0.138} & \textbf{0.032} \\
\bottomrule

\end{tabular}
\caption{\label{tab:ablation}\textbf{Ablation study of AdaBEV on the Air-Co-Pred} $\mathrm{val}$ \textbf{set.} All models are trained using a ResNet-50 backbone and a BEV resolution of $50\times50$. }
\end{table}

\subsection{Detection Decoder and Loss Function}
\label{sec:3.4}

\quad Our method adopts the detection decoder of BEVFormer~\cite{li2022bevformer} for end-to-end prediction. In terms of loss design, we build on the original BEVFormer detection loss $\mathcal{L}_{base}$ by incorporating two auxiliary supervision signals: (1) a perspective loss $\mathcal{L}_{PAS}$ from the PAS module and (2) a contrastive learning loss $\mathcal{L}_{IBCL}$ in the BEV space. The overall loss formulation is defined as follows:
\begin{equation}
\mathcal{L} = \mathcal{L}_{base} + \lambda_{1} \cdot \mathcal{L}_{PAS} + \lambda_{2} \cdot \mathcal{L}_{IBCL}
\end{equation}

\section{Experiments}

\subsection{Experimental Settings}
\label{sec:4.1}

\subsubsection{Dataset and Metrics.}

We evaluate the proposed method on the Air-Co-Pred dataset~\cite{wang2024drones}, a simulated multi-UAV collaborative perception benchmark comprising 200 diverse scenarios. In each scenario, UAVs operate at an altitude of 50 meters, covering an area of approximately $100m\times100m$. Keyframes are annotated at a frequency of 2Hz, with each keyframe containing synchronized RGB images captured from four UAVs. Air-Co-Pred provides a comprehensive suite of metrics for evaluating 3D object detection performance, including Average Precision (AP), Average Translation Error (ATE), Average Scale Error (ASE), and Average Orientation Error (AOE).

\subsubsection{Implementation Details.}

We adopt BEVFormer~\cite{li2022bevformer} as our baseline framework and integrate DD3D~\cite{park2021pseudo} as the detection head for the perspective-aware supervision. We experiment with both ResNet-50 and ResNet-101~\cite{he2016deep}, initialized from ImageNet-pretrained checkpoints that are further fine-tuned on the COCO dataset~\cite{lin2014microsoft} for 2D detection. Apart from our proposed modifications, we follow the default configurations of BEVFormer. In BG-RM, we set the refinement resolution of foreground BEV grids to $r=4$. For IBCL, the number of background feature samples is fixed at 200, with a minimum sampling distance of 4 meters. The BEV representation spans a perception range of $[-51.2m, 51.2m]$ along both the $X$ and $Y$ axes, and a vertical sampling range of $[-53m, -47m]$ along the $Z$ axis. The loss weights for the perspective-aware supervision and the instance-background contrastive learning are set to $\lambda_1=1.0$ and $\lambda_2=2.0$, respectively. We train our models using AdamW optimizer~\cite{loshchilov2017decoupled} with a learning rate of $2\times10^{-4}$. All models are trained for 48 epochs on four NVIDIA RTX 4090 GPUs.

\subsection{3D Object Detection Results}
\label{sec:4.2}

\quad We compare our proposed AdaBEV with several current state-of-the-art BEV detectors. Among them, UCDNet~\cite{tian2024ucdnet} is specifically designed for multi-UAV collaborative object detection, while the others are mainstream 3D detection methods originally developed for autonomous driving scenarios~\cite{caesar2020nuscenes}. We report the 3D object detection results on Air-Co-Pred val set in Table~\ref{tab:main}. 

When using ResNet-50 and a low BEV resolution of $50\times50$, AdaBEV achieves an AP of 0.783, surpassing BEVFormer (0.759) while reducing localization and orientation errors. Remarkably, our method even outperforms several methods with higher BEV resolution and larger model capacity. In particular, AdaBEV surpasses UCDNet, which is specifically designed for multi-UAV collaborative detection with inter-drone information sharing, demonstrating stronger representation and robustness even without explicit collaboration. With ResNet-101 and a high BEV resolution of $150\times150$, AdaBEV achieves the best overall performance. We argue that performing finer-grained refinement beyond a BEV resolution of $200\times200$ is unnecessary, as it introduces substantial computational overhead without yielding meaningful performance gains. We will further elaborate in Section~\ref{sec:4.3} how AdaBEV achieves substantial accuracy improvements with minimal computational overhead, striking a favorable balance between performance and computation complexity.

\subsection{Minimal Overhead, Maximum Improvement}
\label{sec:4.3}

\begin{figure*}[t]
\centering
\includegraphics[width=0.9\textwidth]{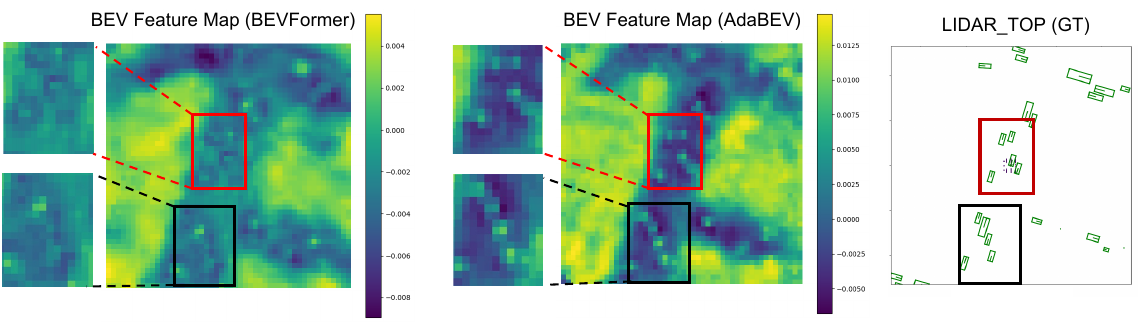} 
\caption{\textbf{Visualization of BEV feature maps in BEVFormer and AdaBEV.} Compared to the BEVFormer, AdaBEV produces more detailed and semantically discriminative BEV features in instance regions. The highlighted regions show clearer object boundaries and improved separability from background and nearby instances.}
\label{fig4}
\end{figure*}
\begin{figure*}[t]
\centering
\includegraphics[width=1.0\textwidth]{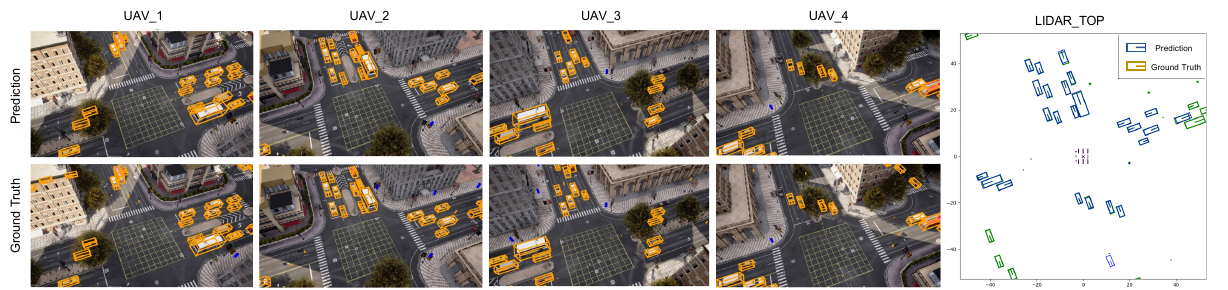} 
\caption{\textbf{Visualization results of AdaBEV on the Air-Co-Pred} $\mathrm{val}$ \textbf{set.} We show the 3D bboxes predictions in multi-UAV images and the bird's-eye view.}
\label{fig5}
\end{figure*}

\quad High-performance 3D detection on resource-constrained UAVs remains challenging, as increasing BEV resolution improves accuracy but incurs significant computational and memory overhead. For instance, upgrading the BEV resolution from $50\times50$ to $200\times200$ in BEVFormer leads to about $2.5\times$ increases in GFLOPs across model scales, rendering high-resolution settings impractical for real-time edge deployment. This motivates the need for methods that achieve near-upper-bound accuracy while maintaining the low-resolution configurations.

To evaluate the effectiveness of AdaBEV under this stringent trade-off, we compare it with the baseline BEVFormer models at both the lower bound ($50\times50$ BEV) and upper bound ($200\times200$ BEV) resolutions, across tiny, small, and base variants. As shown in Table~\ref{tab:up&low}, our method consistently improves the detection accuracy with minimal additional computational cost. On the tiny model, our approach achieves an AP of 0.775, which is remarkably close to the upper bound performance of 0.786 obtained with a $200\times200$ BEV resolution. In particular, this improvement is achieved with only a negligible increase in GFLOPs (141.56 $\mathrm{vs.}$ 141.49), and a modest parameter overhead (41.01M $\mathrm{vs.}$ 33.57M). On the base model, our method attains an AP of 0.934, nearly matching the upper bound result of 0.940, while keeping the computation virtually unchanged (886.32 $\mathrm{vs.}$ 880.67 GFLOPs). In general, our method achieves over 50\% of the AP improvement typically brought by high-resolution BEV settings, with less than 1.5\% additional computation compared to the low-resolution baseline, showing that it is highly suitable for real-world deployment on lightweight autonomous aerial platforms.

\subsection{UAV Collaboration and Ablation Studies}
\label{sec:4.4}

\quad We analyze the effect of the UAV collaboration scale on 3D detection performance on AdaBEV. As shown in Table~\ref{tab:collab}, increasing the number of collaborative UAVs leads to consistent improvements across all metrics. In particular, using four UAVs achieves a relative improvement in AP of 31.6\% over the best two-UAV configuration (0.783 $\mathrm{vs.}$ 0.595), demonstrating the benefits of wider spatial coverage and multi-view fusion. Moreover, the diagonal UAV pair (UAV\textsubscript{13}) slightly outperforms the adjacent pair (UAV\textsubscript{12}), indicating the importance of complementary viewpoints in collaborative perception.

In the ablation study, we progressively ablate the bells and whistles introduced in AdaBEV to confirm the contribution of each proposed component to the final performance. As shown in Table~\ref{tab:ablation}, introducing PAS alone improves the AP from the BEVFormer baseline to 0.764. This demonstrates the positive supervisory effect of incorporating perspective-aware information into BEV representation learning. Adding BG-RM on top of PAS further increases performance to 0.775, a +2.1\% improvement over the baseline, validating the effectiveness of the foreground-guided resolution refinement mechanism in enhancing fine-grained spatial encoding. Finally, integrating all three modules leads to the best overall performance, with an AP of 0.783, representing a +2.4 point improvement (+3.2\%) over the baseline.

\subsection{Visualization}
\label{sec:4.5}

\quad We visualize BEV feature maps and 3D detection results to qualitatively assess the effectiveness of AdaBEV. As shown in Figure~\ref{fig4}, compared to BEVFormer, our method generates more structured and semantically focused BEV features. Object regions exhibit distinguishable boundaries, enhanced contrast against background, and better inter-instance separation, demonstrating the benefits of our adaptive refinement and instance-aware supervision in handling non-uniform semantic distributions typical in UAV scenarios. Figure~\ref{fig5} further presents that AdaBEV shows strong geometric consistency with ground truth in both BEV and image views, even under occlusion or large perspective changes. The predictions remain consistent across all viewpoints, showing strong spatial awareness of the model.

\section{Conclusion}

\quad In this paper, we propose AdaBEV, a refine-and-contrast framework for multi-UAV collaborative 3D object detection that learns adaptive instance-aware BEV representations. To overcome the limitations of uniform BEV modeling in aerial scenarios, our approach introduces a box-guided refinement module (BG-RM) to adaptively enhance foreground BEV grids and an instance-background contrastive learning (IBCL) to strengthen instance awareness in the BEV space. Leveraging 2D supervision and contrastive objectives, our method achieves fine-grained and discriminative BEV representations while maintaining negligible computational overhead. Extensive experiments on the Air-Co-Pred benchmark demonstrate the effectiveness of our framework, setting a new direction for instance-aware, resource-efficient multi-UAV 3D detection tasks.

\bigskip

\bibliography{main}

\begin{thebibliography}{30}
\providecommand{\natexlab}[1]{#1}

\bibitem[{Caesar et~al.(2020)Caesar, Bankiti, Lang, Vora, Liong, Xu, Krishnan, Pan, Baldan, and Beijbom}]{caesar2020nuscenes}
Caesar, H.; Bankiti, V.; Lang, A.~H.; Vora, S.; Liong, V.~E.; Xu, Q.; Krishnan, A.; Pan, Y.; Baldan, G.; and Beijbom, O. 2020.
\newblock nuscenes: A multimodal dataset for autonomous driving.
\newblock In \emph{Proceedings of the IEEE/CVF conference on computer vision and pattern recognition}, 11621--11631.

\bibitem[{Chang et~al.(2018)Chang, Yang, Wu, Shi, and Shi}]{chang2018surveillance}
Chang, X.; Yang, C.; Wu, J.; Shi, X.; and Shi, Z. 2018.
\newblock A surveillance system for drone localization and tracking using acoustic arrays.
\newblock In \emph{2018 IEEE 10th Sensor Array and Multichannel Signal Processing Workshop (SAM)}, 573--577. IEEE.

\bibitem[{Chen et~al.(2020)Chen, Kornblith, Norouzi, and Hinton}]{chen2020simple}
Chen, T.; Kornblith, S.; Norouzi, M.; and Hinton, G. 2020.
\newblock A simple framework for contrastive learning of visual representations.
\newblock In \emph{International conference on machine learning}, 1597--1607. PmLR.

\bibitem[{He et~al.(2016)He, Zhang, Ren, and Sun}]{he2016deep}
He, K.; Zhang, X.; Ren, S.; and Sun, J. 2016.
\newblock Deep residual learning for image recognition.
\newblock In \emph{Proceedings of the IEEE conference on computer vision and pattern recognition}, 770--778.

\bibitem[{Hu et~al.(2022)Hu, Fang, Lei, Zhong, and Chen}]{hu2022where2comm}
Hu, Y.; Fang, S.; Lei, Z.; Zhong, Y.; and Chen, S. 2022.
\newblock Where2comm: Communication-efficient collaborative perception via spatial confidence maps.
\newblock \emph{Advances in neural information processing systems}, 35: 4874--4886.

\bibitem[{Huang and Huang(2022)}]{huang2022bevdet4d}
Huang, J.; and Huang, G. 2022.
\newblock Bevdet4d: Exploit temporal cues in multi-camera 3d object detection.
\newblock \emph{arXiv preprint arXiv:2203.17054}.

\bibitem[{Huang et~al.(2021)Huang, Huang, Zhu, Ye, and Du}]{huang2021bevdet}
Huang, J.; Huang, G.; Zhu, Z.; Ye, Y.; and Du, D. 2021.
\newblock Bevdet: High-performance multi-camera 3d object detection in bird-eye-view.
\newblock \emph{arXiv preprint arXiv:2112.11790}.

\bibitem[{Jiang et~al.(2024)Jiang, Li, Liu, Wang, Jia, Wang, Han, and Zhang}]{jiang2024far3d}
Jiang, X.; Li, S.; Liu, Y.; Wang, S.; Jia, F.; Wang, T.; Han, L.; and Zhang, X. 2024.
\newblock Far3d: Expanding the horizon for surround-view 3d object detection.
\newblock In \emph{Proceedings of the AAAI conference on artificial intelligence}, volume~38, 2561--2569.

\bibitem[{Li et~al.(2023{\natexlab{a}})Li, Ge, Yu, Yang, Wang, Shi, Sun, and Li}]{li2023bevdepth}
Li, Y.; Ge, Z.; Yu, G.; Yang, J.; Wang, Z.; Shi, Y.; Sun, J.; and Li, Z. 2023{\natexlab{a}}.
\newblock Bevdepth: Acquisition of reliable depth for multi-view 3d object detection.
\newblock In \emph{Proceedings of the AAAI conference on artificial intelligence}, volume~37, 1477--1485.

\bibitem[{Li et~al.(2022)Li, Wang, Li, Xie, Sima, Lu, Qiao, and Dai}]{li2022bevformer}
Li, Z.; Wang, W.; Li, H.; Xie, E.; Sima, C.; Lu, T.; Qiao, Y.; and Dai, J. 2022.
\newblock BEVFormer: Learning Bird’s-Eye-View Representation from Multi-camera Images via Spatiotemporal Transformers.
\newblock In \emph{European Conference on Computer Vision}, 1--18.

\bibitem[{Li et~al.(2023{\natexlab{b}})Li, Yu, Wang, Anandkumar, Lu, and Alvarez}]{li2023fb}
Li, Z.; Yu, Z.; Wang, W.; Anandkumar, A.; Lu, T.; and Alvarez, J.~M. 2023{\natexlab{b}}.
\newblock Fb-bev: Bev representation from forward-backward view transformations.
\newblock In \emph{Proceedings of the IEEE/CVF International Conference on Computer Vision}, 6919--6928.

\bibitem[{Lin et~al.(2014)Lin, Maire, Belongie, Hays, Perona, Ramanan, Doll{\'a}r, and Zitnick}]{lin2014microsoft}
Lin, T.-Y.; Maire, M.; Belongie, S.; Hays, J.; Perona, P.; Ramanan, D.; Doll{\'a}r, P.; and Zitnick, C.~L. 2014.
\newblock Microsoft coco: Common objects in context.
\newblock In \emph{European conference on computer vision}, 740--755. Springer.

\bibitem[{Liu et~al.(2022)Liu, Wang, Zhang, and Sun}]{liu2022petr}
Liu, Y.; Wang, T.; Zhang, X.; and Sun, J. 2022.
\newblock Petr: Position embedding transformation for multi-view 3d object detection.
\newblock In \emph{European conference on computer vision}, 531--548. Springer.

\bibitem[{Loshchilov and Hutter(2017)}]{loshchilov2017decoupled}
Loshchilov, I.; and Hutter, F. 2017.
\newblock Decoupled weight decay regularization.
\newblock \emph{arXiv preprint arXiv:1711.05101}.

\bibitem[{Nguyen and Nguyen(2021)}]{nguyen2021drone}
Nguyen, H. P.~D.; and Nguyen, D.~D. 2021.
\newblock Drone application in smart cities: The general overview of security vulnerabilities and countermeasures for data communication.
\newblock \emph{Development and Future of Internet of Drones (IoD): Insights, Trends and Road Ahead}, 185--210.

\bibitem[{Oord, Li, and Vinyals(2018)}]{oord2018representation}
Oord, A. v.~d.; Li, Y.; and Vinyals, O. 2018.
\newblock Representation learning with contrastive predictive coding.
\newblock \emph{arXiv preprint arXiv:1807.03748}.

\bibitem[{Pan et~al.(2024)Pan, Yaman, Velipasalar, and Ren}]{pan2024clip}
Pan, C.; Yaman, B.; Velipasalar, S.; and Ren, L. 2024.
\newblock Clip-bevformer: Enhancing multi-view image-based bev detector with ground truth flow.
\newblock In \emph{Proceedings of the IEEE/CVF Conference on Computer Vision and Pattern Recognition}, 15216--15225.

\bibitem[{Park et~al.(2021)Park, Ambrus, Guizilini, Li, and Gaidon}]{park2021pseudo}
Park, D.; Ambrus, R.; Guizilini, V.; Li, J.; and Gaidon, A. 2021.
\newblock Is pseudo-lidar needed for monocular 3d object detection?
\newblock In \emph{Proceedings of the IEEE/CVF international conference on computer vision}, 3142--3152.

\bibitem[{Philion and Fidler(2020)}]{philion2020lift}
Philion, J.; and Fidler, S. 2020.
\newblock Lift, splat, shoot: Encoding images from arbitrary camera rigs by implicitly unprojecting to 3d.
\newblock In \emph{European conference on computer vision}, 194--210. Springer.

\bibitem[{Qu et~al.(2023)Qu, Sorbelli, Singh, Calyam, and Das}]{qu2023environmentally}
Qu, C.; Sorbelli, F.~B.; Singh, R.; Calyam, P.; and Das, S.~K. 2023.
\newblock Environmentally-aware and energy-efficient multi-drone coordination and networking for disaster response.
\newblock \emph{IEEE transactions on network and service management}, 20(2): 1093--1109.

\bibitem[{Tian et~al.(2024)Tian, Wang, Cheng, Wang, Wang, Zhao, Yan, Yang, and Sun}]{tian2024ucdnet}
Tian, P.; Wang, Z.; Cheng, P.; Wang, Y.; Wang, Z.; Zhao, L.; Yan, M.; Yang, X.; and Sun, X. 2024.
\newblock Ucdnet: Multi-uav collaborative 3d object detection network by reliable feature mapping.
\newblock \emph{IEEE Transactions on Geoscience and Remote Sensing}.

\bibitem[{Vaswani et~al.(2017)Vaswani, Shazeer, Parmar, Uszkoreit, Jones, Gomez, Kaiser, and Polosukhin}]{vaswani2017attention}
Vaswani, A.; Shazeer, N.; Parmar, N.; Uszkoreit, J.; Jones, L.; Gomez, A.~N.; Kaiser, {\L}.; and Polosukhin, I. 2017.
\newblock Attention is all you need.
\newblock \emph{Advances in neural information processing systems}, 30.

\bibitem[{Wang, Jiang, and Li(2023)}]{wang2023focal}
Wang, S.; Jiang, X.; and Li, Y. 2023.
\newblock Focal-petr: Embracing foreground for efficient multi-camera 3d object detection.
\newblock \emph{IEEE Transactions on Intelligent Vehicles}, 9(1): 1481--1489.

\bibitem[{Wang et~al.(2024)Wang, Cheng, Chen, Tian, Wang, Li, Yang, and Sun}]{wang2024drones}
Wang, Z.; Cheng, P.; Chen, M.; Tian, P.; Wang, Z.; Li, X.; Yang, X.; and Sun, X. 2024.
\newblock Drones help drones: A collaborative framework for multi-drone object trajectory prediction and beyond.
\newblock \emph{Advances in Neural Information Processing Systems}, 37: 64604--64628.

\bibitem[{Wang et~al.(2023)Wang, Huang, Fu, Wang, and Liu}]{wang2023object}
Wang, Z.; Huang, Z.; Fu, J.; Wang, N.; and Liu, S. 2023.
\newblock Object as query: Lifting any 2d object detector to 3d detection.
\newblock In \emph{Proceedings of the IEEE/CVF International Conference on Computer Vision}, 3791--3800.

\bibitem[{Yang et~al.(2023)Yang, Chen, Tian, Tao, Zhu, Zhang, Huang, Li, Qiao, Lu et~al.}]{yang2023bevformer}
Yang, C.; Chen, Y.; Tian, H.; Tao, C.; Zhu, X.; Zhang, Z.; Huang, G.; Li, H.; Qiao, Y.; Lu, L.; et~al. 2023.
\newblock Bevformer v2: Adapting modern image backbones to bird's-eye-view recognition via perspective supervision.
\newblock In \emph{Proceedings of the IEEE/CVF conference on computer vision and pattern recognition}, 17830--17839.

\bibitem[{Yang et~al.(2024)Yang, Lin, Huang, and Crowley}]{yang2024widthformer}
Yang, C.; Lin, T.; Huang, L.; and Crowley, E.~J. 2024.
\newblock Widthformer: Toward efficient transformer-based bev view transformation.
\newblock In \emph{2024 IEEE/RSJ International Conference on Intelligent Robots and Systems (IROS)}, 8457--8464. IEEE.

\bibitem[{Zhou and Kr{\"a}henb{\"u}hl(2022)}]{zhou2022cross}
Zhou, B.; and Kr{\"a}henb{\"u}hl, P. 2022.
\newblock Cross-view transformers for real-time map-view semantic segmentation.
\newblock In \emph{Proceedings of the IEEE/CVF conference on computer vision and pattern recognition}, 13760--13769.

\bibitem[{Zhu et~al.(2020{\natexlab{a}})Zhu, Zheng, Du, Wen, Sun, and Hu}]{zhu2020multi}
Zhu, P.; Zheng, J.; Du, D.; Wen, L.; Sun, Y.; and Hu, Q. 2020{\natexlab{a}}.
\newblock Multi-drone-based single object tracking with agent sharing network.
\newblock \emph{IEEE Transactions on Circuits and Systems for Video Technology}, 31(10): 4058--4070.

\bibitem[{Zhu et~al.(2020{\natexlab{b}})Zhu, Su, Lu, Li, Wang, and Dai}]{zhu2020deformable}
Zhu, X.; Su, W.; Lu, L.; Li, B.; Wang, X.; and Dai, J. 2020{\natexlab{b}}.
\newblock Deformable detr: Deformable transformers for end-to-end object detection.
\newblock \emph{arXiv preprint arXiv:2010.04159}.

\end{thebibliography}

\end{document}